\documentclass{article}
\usepackage{iclr2027_conference,times}
\usepackage[T1]{fontenc}
\usepackage[utf8]{inputenc}
\usepackage{amsmath,amssymb,graphicx,booktabs,tabularx,longtable,array}
\usepackage{xcolor,hyperref,url,needspace}
\graphicspath{{figures/}}

\newif\ifanonymous
\anonymousfalse
\ifanonymous\else\iclrfinalcopy\fi
\definecolor{RDInk}{HTML}{1D1D1F}
\definecolor{RDMuted}{HTML}{5F5F68}
\definecolor{RDAccent}{HTML}{AC9FE3}
\definecolor{RDAccentInk}{HTML}{6A5CC0}
\definecolor{RDAccentSoft}{HTML}{F3F0FC}
\definecolor{RDLine}{HTML}{D7D4E2}

\ifanonymous
  \hypersetup{colorlinks=true,allcolors=black}
  \newcommand{\RDTitle}[1]{#1}
  \newcommand{\RDCaptionTitle}[1]{\textbf{#1}}
  \newcommand{\RDTableHeading}[1]{\textbf{#1}}
  \newcommand{\RDTableEmphasis}[1]{\textbf{#1}}
\else
  \hypersetup{colorlinks=true,linkcolor=RDAccentInk,
    citecolor=RDAccentInk,urlcolor=RDAccentInk}
  \newcommand{\RDTitle}[1]{\textcolor{RDAccentInk}{#1}}
  \newcommand{\RDCaptionTitle}[1]{\textcolor{RDAccentInk}{\textbf{#1}}}
  \newcommand{\RDTableHeading}[1]{\textcolor{RDAccentInk}{\textbf{#1}}}
  \newcommand{\RDTableEmphasis}[1]{\textcolor{RDAccentInk}{\textbf{#1}}}
  \AtBeginDocument{\color{RDInk}}

  \makeatletter
  \let\RDOriginalStartSection\@startsection
  \def\@startsection#1#2#3#4#5#6{%
    \RDOriginalStartSection{#1}{#2}{#3}{#4}{#5}%
      {#6\ifnum#2=1\relax\color{RDAccentInk}\else\color{RDInk}\fi}}
  \let\RDOriginalMakeCaption\@makecaption
  \long\def\@makecaption#1#2{%
    \RDOriginalMakeCaption{\textcolor{RDAccentInk}{#1}}{#2}}
  \makeatother
  \let\RDOriginalHeadRule\headrule
  \renewcommand{\headrule}{{\color{RDLine}\RDOriginalHeadRule}}
\fi

\newcommand{\RDTableSetup}{%
  \normalsize
  \renewcommand{\arraystretch}{1.3}%
  \setlength{\tabcolsep}{4pt}%
  \setlength{\abovetopsep}{5pt}%
  \setlength{\belowrulesep}{3pt}%
  \setlength{\aboverulesep}{3pt}}
\newcolumntype{L}[1]{>{\raggedright\arraybackslash}p{#1}}
\newcolumntype{R}[1]{>{\raggedleft\arraybackslash}p{#1}}
\newcolumntype{Y}{>{\raggedright\arraybackslash}X}
\newlength{\RDTableContentWidth}

\newcommand{\method}{\textbf{RoboDoctor}}
\newcommand{\Hybrid}{\textsc{Hybrid}}
\newcommand{\Direct}{\textsc{Direct}}
\newcommand{\code}[1]{\texttt{#1}}
\newcommand{\R}{\mathbb{R}}

\title{\RDTitle{RoboDoctor}: Training-Free Recursive\newline Self-Improvement for Robot Policies}
\author{
\small \textbf{Heyang Yu$^{1,*}$, Haomin Li$^{1,*}$, Sihang Li$^{2}$, Jixin Yan$^{3}$, Shiyun Peng$^{1}$,}\\
\small \textbf{Wanqi Zhong$^{1}$, Tianyu Xin$^{1}$, Yuan Yao$^{1}$, Xiaoyu Tian$^{1,\dagger}$, Yiming Li$^{1,\dagger}$}\\[3pt]
\small $^{1}$Tsinghua University\quad
$^{2}$New York University\quad
$^{3}$Northwestern University\\
\small $^{*}$Equal contribution\qquad
$^{\dagger}$Equal correspondence
}

\newcommand{\WarmupScore}{49.74}
\newcommand{\RefineScore}{57.10}
\newcommand{\ScoreGain}{7.37}
\newcommand{\WarmupSuccess}{40.83}
\newcommand{\RefineSuccess}{49.44}
\newcommand{\SuccessGain}{8.61}

\begin{document}
\maketitle
\ifanonymous\else\lhead{\textcolor{RDMuted}{RoboDoctor Technical Report}}\fi

\begin{abstract}
Generalist robot policies and language-model agents provide complementary capabilities, but completing a task also requires adapting their use to the consequences of physical interaction. We study execution as a locus of adaptation: feedback guides the current action and supplies evidence for revising subsequent decisions under fixed model weights. We introduce \method{}, a training-free framework that connects diagnosis, grounded action, and persistent experience across three revision targets. The \emph{harness} allocates control between policy and agent; \emph{intervention} determines when to assist while preserving useful policy behavior; and \emph{operation} refines physical methods through observed effects. Linked experience records retain the reason for a decision, the attempted method, its outcome, and conditions for reuse. On a curated archive of 54 RoboDojo-Sim task entries, RoboDoctor achieves a five-dimension mean score of \RefineScore{}, with a native success rate of \RefineSuccess\%. Recorded interactions illustrate strategy changes, selective correction, and grasp revision. These findings characterize execution-time adaptation in the evaluated system; controlled trials remain necessary to isolate the contribution of cross-episode experience.
\end{abstract}
\begin{figure}[htbp]
  \centering
  \includegraphics[width=0.85\linewidth]{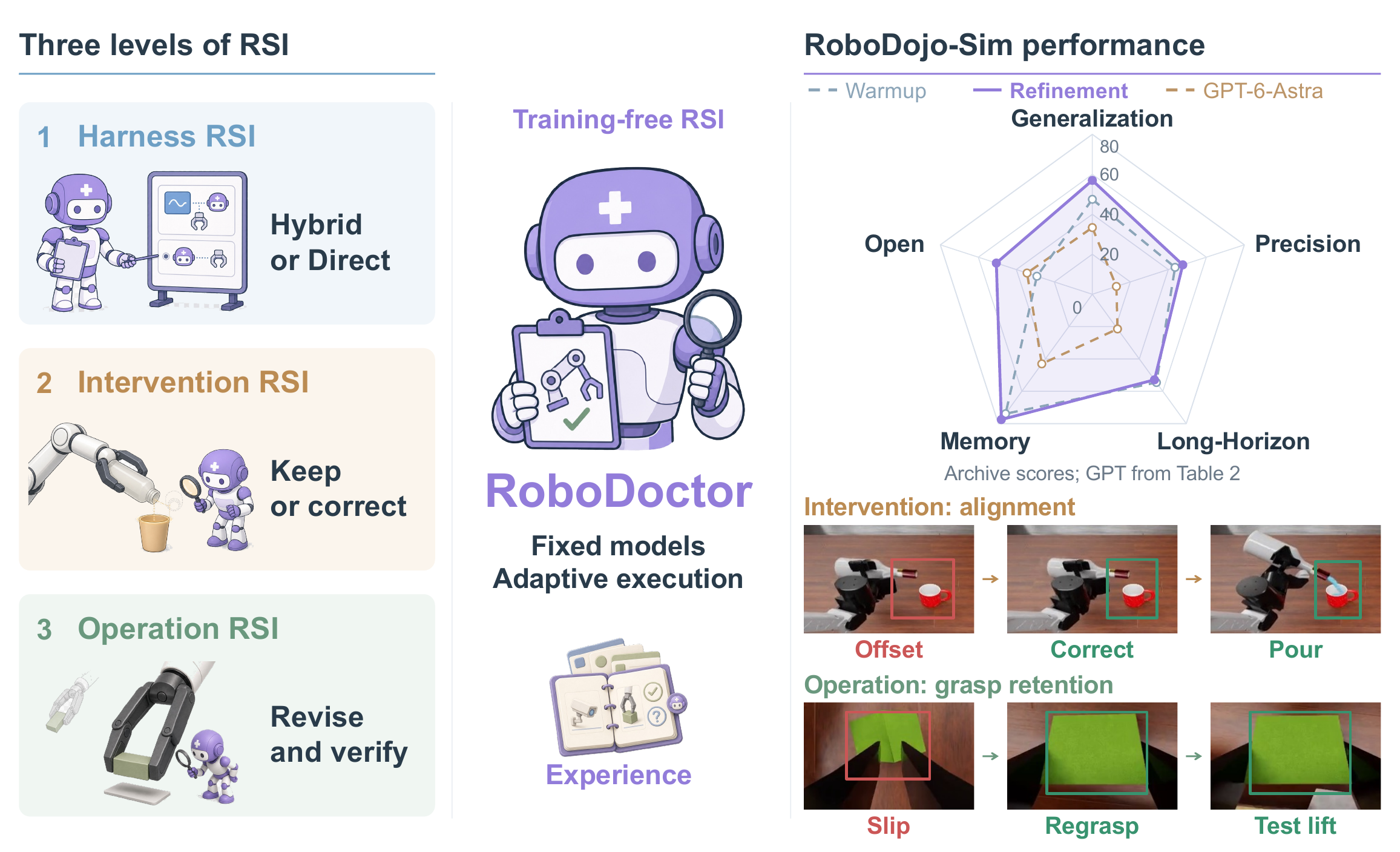}
  \caption{\RDCaptionTitle{RoboDoctor: training-free adaptation across three levels of execution.} Harness RSI allocates control between the policy and the Doctor; Intervention RSI decides when to retain or correct policy motion; Operation RSI revises physical methods and verifies their effects. External experience connects these decisions while both models remain frozen. The radar chart summarizes five capability scores for Warmup, Refinement, and the published GPT-6-Astra reference. Recorded examples illustrate alignment correction followed by pouring, and grasp revision followed by a test lift.}

  \label{fig:overview}
\end{figure}
\section{Introduction}
Vision--language--action (VLA) models  and world action models (WAM) have expanded the range of manipulation behaviors available from a single model \citep{kim2024openvla,black2025pi05,ye2026dreamzero}. Language-model agents add complementary capabilities in task interpretation, planning, and tool use \citep{ahn2022saycan,liang2022code}. Translating these capabilities into complete task execution also requires managing their consequences: each interaction changes the conditions under which later actions must succeed. A grasp can be locally successful yet leave an object poorly positioned for the next step. Robot capability therefore depends on how available behaviors are coordinated, evaluated, and revised through physical interaction. This raises a system-level question: \emph{how can a robot use execution feedback to adapt the way it deploys existing model capabilities?}

Robot agents already incorporate environmental feedback into planning, detect execution failures, and use reflection to refine action selection \citep{huang2022innermonologue,guo2024doremi,feng2025reflective}. Building on this progress, we study the decisions that arise when an agent and a continuous robot policy share responsibility for execution. A deviation may require a local correction, a different contact method, or a change in the strategy that organizes the task. These responses operate at different levels and can interfere with one another: correcting alignment may preserve a useful grasp, whereas repeatedly restoring that grasp may perpetuate an ineffective strategy. The challenge is to identify what needs revision while retaining behavior that remains useful.

Recent embodied recursive self-improvement (RSI) systems use physical feedback to revise execution harnesses and accumulate reusable skills \citep{physicalrsi2026,song2026embodiedrsi,li2026embodiedevo}. Within this direction, our perspective is that \emph{execution is itself a locus of adaptation}. Physical interaction serves two roles: it advances the task and provides evidence about how the available capabilities should be used. That evidence can regulate the next action and become external experience for subsequent attempts. With both policy and agent weights fixed, the adaptive state lies in control organization, intervention decisions, and the accumulated methods that inform future behavior. Preserving successful behavior is central to this process: a failure should prompt revision at the level implicated by the feedback, without discarding competence that the same interaction has demonstrated.

Making this perspective operational requires connecting diagnosis to physical evidence. Recognizing a slipped grasp does not specify a stable contact, and reaching a commanded pose does not establish that the object moved as intended. Likewise, a successful correction supports a reusable lesson only under conditions justified by the observation. An alignment principle may apply in another scene even when its original displacement does not. We therefore link the reason for a decision to the attempted physical method, its observed effect, and its conditions of applicability. Reuse calls for checking those conditions and grounding the method again in the current scene.

We introduce \method{}, a framework that implements this perspective through three coupled revision targets (Figure~\ref{fig:overview}). \emph{Harness RSI} allocates control responsibility between policy and agent. \emph{Intervention RSI} determines when assistance is warranted and which policy behavior to preserve. \emph{Operation RSI} refines how an intended physical change is achieved. A \Hybrid{} controller lets the policy propose primary motion while the Doctor reviews and selectively corrects it; a \Direct{} controller assigns planning and action to the Doctor through the same grounded tools \citep{su2026astra}. Observed outcomes guide selection between these controllers, and linked experience records support revision within and across episodes. Recursive self-improvement here refers to updating and reusing this external state under frozen model weights; the evaluated harness selects between predefined controllers.

We evaluate \method{} through a curated archive of 54 RoboDojo-Sim task entries \citep{chen2026robodojo}. The selected Refinement outcomes achieve an Overall score of \RefineScore{}, compared with \WarmupScore{} during Warmup. The capability profile in Figure~\ref{fig:overview} shows that the largest gains occur in Generalization and Open instruction following, while Long-Horizon performance decreases. The accompanying examples illustrate alignment correction and grasp revision; Section~\ref{sec:cases} examines these interactions alongside a change in task strategy. The archive supports analysis of these mechanisms and their associated performance; its retrospective selection and unequal coverage do not isolate the causal benefit of cross-episode experience (Section~\ref{sec:protocol}).

Our contributions are: (i) a training-free execution framework that connects control allocation, selective intervention, and physical operation as coupled targets of adaptation; (ii) an experience mechanism that links diagnosis, method, outcome, and applicability to guide subsequent use of fixed model capabilities; and (iii) a 54-task empirical analysis combining aggregate outcomes with recorded evidence of how execution decisions change.

\section{Related Work}
\paragraph{VLA and WAM based robot policies.}
Learned robot policies provide the manipulation capabilities on which execution-time adaptation can operate. Diffusion Policy models multimodal action distributions for visuomotor control \citep{chi2023diffusion}. Octo, OpenVLA, and $\pi_0$ develop generalist robot policies \citep{octo2024octo,kim2024openvla,black2025pi0}, while $\pi_{0.5}$ uses heterogeneous co-training to extend generalization to new environments \citep{black2025pi05}. World action models connect action generation to prediction of physical evolution: DreamZero jointly models future video and actions \citep{ye2026dreamzero}, and PrismWAM separates semantic context from depth-based physical state \citep{xin2026prismwam}. We study adaptation in how a trained policy is used, evaluating frozen PrismWAM; transfer to other policy families remains unmeasured.

\paragraph{Robot agents and execution-time reasoning.}
Robot agents organize model capabilities into task-directed behavior. SayCan, Code as Policies, and VoxPoser connect language reasoning to robotic affordances, programs, and spatial action representations \citep{ahn2022saycan,liang2022code,huang2023voxposer}. Inner Monologue incorporates environmental feedback into planning \citep{huang2022innermonologue}; REFLECT explains failures from multisensory experience summaries \citep{liu2023reflect}; and DoReMi monitors plan--execution misalignment to support recovery \citep{guo2024doremi}. Reflective Planning uses imagined future states and reflection to refine VLM action selection at test time \citep{feng2025reflective}. We build on feedback-driven execution to study assistance to an existing continuous policy: which behavior to preserve, when to replace a proposal, and whether the correction addresses the task's limitation.

\paragraph{Recursive self-improvement for robots.}
Embodied RSI extends execution-time reasoning by revising the system that generates subsequent behavior. Its adaptive state can take several forms. PhysicalRSI and RegenHarness revise execution harnesses and the rules connecting context, routing, and recovery \citep{physicalrsi2026,wang2026regenharness}. EmbodiedRSI combines hypothesis-guided experimentation with code--skill co-evolution \citep{song2026embodiedrsi}; Embodied Turing Machines develops a shared stateful code library that executes without a VLM or VLA in the loop \citep{hu2026etm}. EmbodiRSI instead uses failure feedback to target corrective data collection and successive policy training \citep{lang2026embodirsi}. \method{} places adaptation in control allocation and external experience while keeping both model weights fixed. The evaluated harness makes a discrete controller selection; within that controller, accumulated evidence informs intervention decisions and physical methods. This mechanism implements a bounded form of robot RSI.

\paragraph{Experience and in-context adaptation.}
For fixed-weight agents, external experience lets an earlier interaction inform later behavior. ReAct interleaves reasoning and action, Reflexion retains verbal feedback, and Voyager accumulates executable skills \citep{yao2022react,shinn2023reflexion,wang2023voyager}. In robotics, RoboICL studies control through in-context demonstrations and interaction memory \citep{liu2026roboicl}, while GPT-Policy organizes demonstrations and execution feedback into context for VLM-based control \citep{cheng2026incontextrobotlearningvlm}. EmbodiedEvo distills physical feedback into reusable code Skills and diagnostic Lessons, acquired in a task-specific scene and then frozen for execution in new scenes \citep{li2026embodiedevo}. We focus on the physical evidence that makes a lesson applicable: its diagnostic trigger, the policy behavior to retain, the attempted operation, and the observed effect.

% Keep the section title, first subsection and opening text on the same page.
\Needspace{9\baselineskip}
\section{RoboDoctor}
\label{sec:method}
\subsection{Execution-Time Adaptation with Frozen Models}
\begin{figure}[t]
  \centering
  \includegraphics[width=\linewidth]{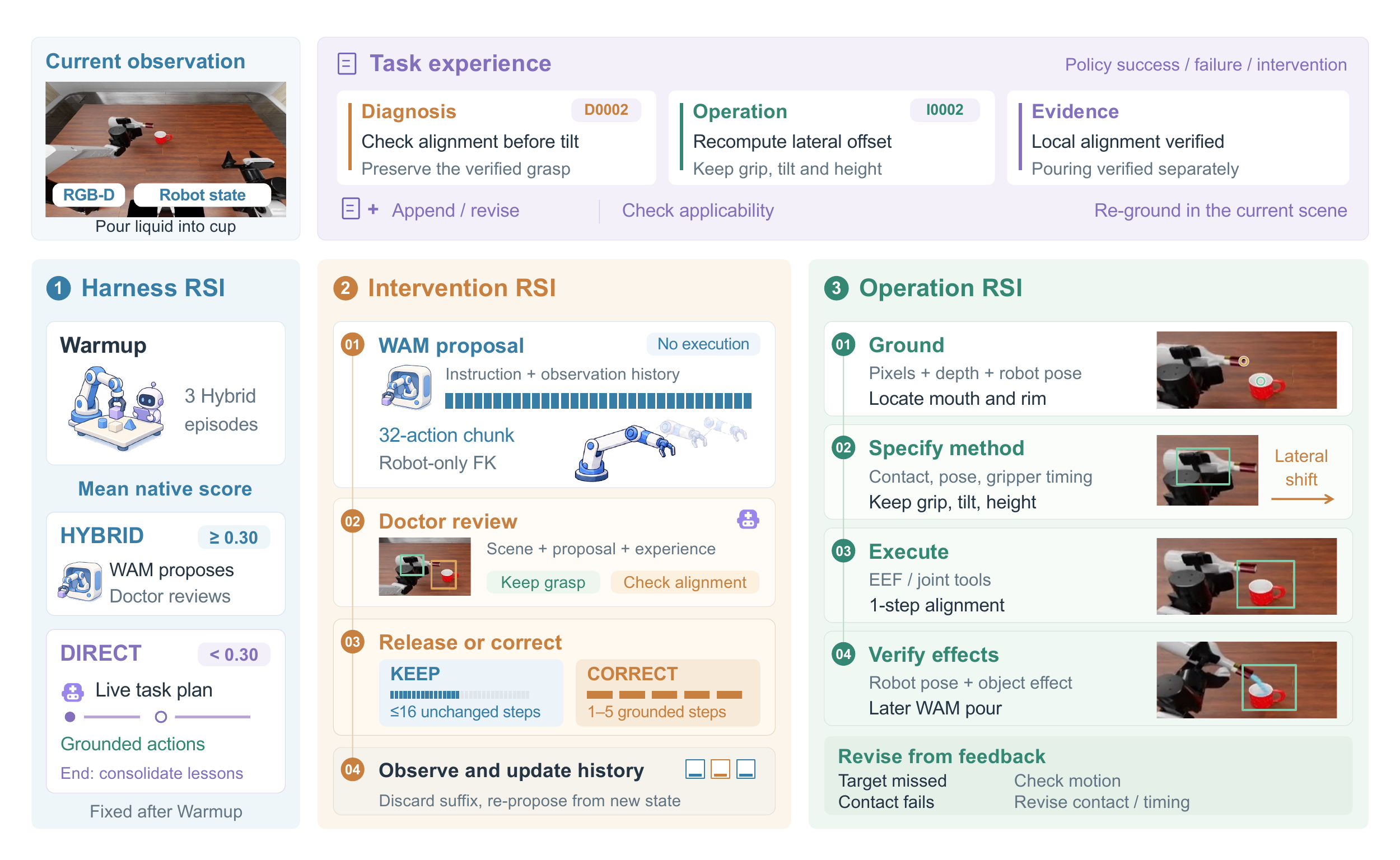}
  \caption{\RDCaptionTitle{RoboDoctor architecture: experience-guided control and grounded execution.} After three HYBRID Warmup episodes, Harness RSI selects HYBRID or DIRECT control and keeps this choice fixed. In HYBRID, the Doctor reviews a WAM proposal and its robot-only forward-kinematics preview, then retains up to 16 unchanged policy steps or issues a grounded correction of 1–5 physical steps. Corrections and DIRECT actions ground the current scene, execute a physical method, and verify its observed effect. The illustrated pouring episode separates a lateral alignment correction that preserves the grasp from subsequent policy-led pouring. Task experience links diagnostic cues, attempted operations, and observed outcomes to inform later decisions.}
  \label{fig:architecture}
\end{figure}

\method{} adapts the use of fixed models through physical feedback. Figure~\ref{fig:architecture} shows how control allocation, selective intervention, and grounded operation connect current observations to experience retrieval and update. Control allocation determines who directs execution, intervention determines what policy behavior to preserve, and operation determines how to achieve and verify a physical effect. Experience links these decisions to their observed consequences and conditions for reuse.

We consider language-specified, dual-arm manipulation. At step $t$, the Doctor receives head and wrist RGB views, metric depth, measured joint state, and end-effector poses, without privileged object poses or object-state labels. A frozen multimodal agent $D_\phi$ acts as the Doctor. A frozen policy $\pi_\theta$ proposes an action chunk from instruction $\ell$ and observation history $h_t$:
\begin{equation}
 A_t=\pi_\theta(h_t,\ell)\in\R^{H\times14}.
\end{equation}
Each action contains six absolute joint targets and one gripper opening per arm; the reported PrismWAM interface uses $H=32$. The adaptive state consists of a selected controller and task-specific external experience. Control allocation is chosen after Warmup and then held fixed for the run. Within the selected mode, fresh observations and revised experience inform subsequent intervention and operation decisions while $\theta$ and $\phi$ remain unchanged.

\subsection{Harness RSI: Allocate Control Responsibility}
\label{sec:harness}
The first decision is which model should direct task execution. Useful local policy motion may support the task, while an ineffective task strategy can require agent-led planning. The two branches in Figure~\ref{fig:architecture} instantiate these alternatives. In \Hybrid{}, the policy proposes primary motion and the Doctor determines what to release or correct. In \Direct{}, the Doctor chooses the interaction strategy and issues all motion through grounded tools. Both modes use the same sensory and actuation interface.

The evaluated harness uses early native outcomes as coarse evidence for this allocation. Each run begins with three \Hybrid{} Warmup episodes, $k=0,1,2$, carrying intervention experience between episodes. Let $\bar{s}_{\mathrm W}=\frac13\sum_{k=0}^{2}s_k$ for valid native scores $s_k\in[0,1]$. The subsequent controller is
\begin{equation}
 c_k=\begin{cases}
 \Direct{}, & \bar{s}_{\mathrm W}<0.30,\\
 \Hybrid{}, & \bar{s}_{\mathrm W}\geq0.30,
 \end{cases}\qquad k\geq3.
 \label{eq:gate}
\end{equation}
Continuing \Hybrid{} preserves the accumulated intervention experience, with harness self-revision based on the Warmup stage. Switching to \Direct{} initializes a separate task-experience document, which subsequent \Direct{} episodes inherit. The selected controller remains fixed; the gate provides a coarse allocation rather than a diagnosis of a specific failure. Subsequent adaptation acts through intervention, operation, and experience.

\subsection{Intervention RSI: Adapt Assistance to Execution Evidence}
\label{sec:intervention}
When the policy supplies primary motion, adaptation requires deciding which behavior to preserve and when assistance is warranted. We separate policy proposal from physical execution: a rollout returns $A_t$ and a robot-only forward-kinematics (FK) preview without moving the robot. The Doctor relates this candidate motion to current RGB-D observations, measured poses, and experience. It seeks an action-relevant reason to assist, such as a changed object location or an impending alignment mismatch, while retaining useful aspects of the interaction.

For each proposal, the Doctor selects one of two execution branches:
\begin{equation}
 a_t=\begin{cases}
 A_t[1{:}m], & \text{release an unchanged prefix},\quad 1\leq m\leq16,\\
 u_t, & \text{substitute a grounded correction},\quad 1\leq b(u_t)\leq5,
 \end{cases}
 \label{eq:control}
\end{equation}
where $b(u_t)$ is the correction's physical step budget. Both branches discard the unused proposal suffix and return a fresh observation. Multi-call recovery is possible, but each correction requires a new observation and policy proposal. Every executed frame, including agent corrections, enters the policy history so that subsequent proposals respond to the achieved state.

The resulting observation tests the intervention decision. Releasing a proposal may expose a missed failure; a correction may restore alignment or disturb a stable grasp. These events can be recorded as evidence for revising later assistance, while autonomous successes identify behavior worth preserving (Section~\ref{sec:experience}). Thus the decision to intervene depends on both the current proposal and experience from earlier outcomes. The FK preview supports reasoning about intended robot motion; the physical effect of either branch must be established through execution.

\subsection{Operation RSI: Ground, Test, and Revise Physical Methods}
\label{sec:operation}
An intervention decision identifies a change to make; operation determines how to realize and verify that change. The same requirement applies when \Direct{} plans a task action. The Operation RSI module in Figure~\ref{fig:architecture} organizes both types of agent-issued action into grounding, action, and verification. The pouring example distinguishes an alignment correction from the subsequent observation of liquid transfer. The Doctor grounds visible surfaces from image pixels and depth, then uses measured end-effector poses and current head or wrist views to construct a contact method. A grounded surface point is a localization reference, not necessarily a valid finger contact or end-effector target. The method specifies approach, contact region, wrist orientation, gripper timing, motion sequence, and the role of the other arm.

Execution provides evidence for retaining or revising this method. The Doctor compares the command, achieved robot motion, and observed object interaction. Failure to reach a target directs attention to motion or configuration; reaching it while the object slips suggests revisiting the contact method. It can revise the relevant choice, for example by changing the gripper closing axis and checking retention with a short lift before transport. End-effector and joint-space tools support these revisions. \Hybrid{} retains the five-step per-call limit; \Direct{} can allocate longer budgets to clear travel and shorter ones to contact-sensitive motion.

Task progress remains subject to the same verification. In \Direct{}, a live plan tracks subtasks and observed progress, while separate notes preserve detailed evidence. Later observations can reopen a subtask if its result is disturbed. In both modes, an operation is retained when its observed effect supports the intended interaction and reconsidered when later evidence contradicts that assessment. Persistent experience carries the method and its supporting evidence into subsequent decisions.

\subsection{Experience: Turn Physical Feedback into Reusable Methods}
\label{sec:experience}
A correction changes the current physical state; retaining its evidence can change how a later action is chosen. Experience provides this link between execution and adaptation. A lesson connects the reason for a decision, the behavior worth preserving, the attempted method, and its observed effect. Its applicability is limited by that evidence: achieving local alignment, for example, does not establish that the subsequent pour succeeded.

In \Hybrid{}, records contain an event type, triggering evidence, policy state, attempted method when applicable, and outcome. Policy state distinguishes observations, proposed or executed actions, and measured or previewed FK. Operation fields describe timing, tools, contact technique, and a stopping condition. Linked diagnosis and operation entries associate the evidence for intervention with the attempted physical method. The experience panel in Figure~\ref{fig:architecture} links a mouth–rim alignment cue to a lateral correction that preserves the grasp, with local alignment and subsequent pouring verified separately. The schema distinguishes autonomous policy success, policy failure, and intervention. A policy-success record specifies the observed object interaction, success evidence, and applicability; other records require the attempted operation and stopping condition. Failed attempts remain evidence about when a method should be revised.

\paragraph{Updating experience from an observed event.}
At an informative event, the Doctor produces a record $r_t$ and either a new identifier or an existing identifier $i$ to revise. Appending retains a new observation and method; replacing revises the account associated with an earlier identifier. Let $\mathcal{V}(r_t)$ denote structural validation and $M_t$ the current record collection. The accepted update is
\begin{equation}
 M_{t+1}=\begin{cases}
   M_t\mathbin{\|}[r_t], & \mathcal{V}(r_t),\quad\text{new identifier},\\
   \operatorname{replace}(M_t,i,r_t), & \mathcal{V}(r_t),\quad\text{revise identifier }i,\\
   M_t, & \text{otherwise},
 \end{cases}
 \label{eq:memory}
\end{equation}
where $\|$ denotes append and $\operatorname{replace}(M_t,i,r_t)$ removes the record with identifier $i$ and appends $r_t$, preserving the relative order of the other records. Accepted updates increment the document version and atomically save the snapshot; a rejected update returns an error. The agent determines the semantic revision. Host validation checks required fields, rather than independently proving the diagnosis or resolving contradictions.

\paragraph{Re-grounding before reuse.}
The updated \Hybrid{} snapshot accompanies subsequent observations and is exported at episode termination. The Doctor consults this document as context, checks applicability, and recomputes action targets from fresh geometry. The retrieval and update paths in Figure~\ref{fig:architecture} connect retained diagnostic cues and physical methods to subsequent execution. For the pouring case in Figure~\ref{fig:cases}(b), the reusable method is to check mouth--rim alignment before deepening a tilt while preserving a stable grasp. The required offset must be recomputed for the new scene, and successful local alignment leaves pouring to be verified separately.

\Direct{} consolidates task lessons after termination in an action-disabled summary turn. It combines applicable prior lessons with new methods, failures, and unresolved questions; the host attaches native validity, score, and success. The next episode reads this experience before planning, establishing current object poses and completed subtasks anew. The loop closes when a retained lesson informs another decision and that decision produces fresh evidence for revision. Both modes therefore support adaptation through external state under fixed model weights, without guaranteeing that each revision improves performance.

% Keep the section title, first subsection and opening text on the same page.
\Needspace{9\baselineskip}
\section{Experimental Evaluation}
\label{sec:experiments}

\subsection{Benchmark and Evaluation Protocol}
\label{sec:protocol}

We evaluate performance on 54 task entries from RoboDojo-Sim across five capability dimensions: Generalization, Precision, Long-Horizon, Memory, and Open instruction following. Generalization comprises 12 standard and 12 randomized entries, while the remaining dimensions comprise 8, 8, 6, and 8 tasks, respectively. Our evaluation uses the category mapping and reference leaderboard archived on 2026-10-20. We use PrismWAM as the underlying policy \citep{xin2026prismwam}. The local archive combines multiple source runs for which a uniform agent version, checkpoint, and inference budget are not established.

\paragraph{Evaluation stages and outcome selection.}
We divide the evaluation into two stages: Warmup, covering iterations 0--2, and Refinement, covering iterations 3--8. Warmup already incorporates Doctor assistance and experience updates and therefore characterizes the initial stage of the system rather than an unassisted policy or a baseline without memory.

An outcome is considered valid if it is complete, explicitly marked as valid for evaluation, and associated with a finite native score in $[0,1]$. Valid outcomes with zero scores are retained. The archive includes score-based selection among candidate outcomes, retaining the highest valid score with deterministic tie-breaking. Because selected outcomes may originate from different source runs and memory histories, the stage comparison characterizes the archived results rather than a prospectively controlled treatment effect.

\paragraph{Evaluation metrics.}
Let $V_{j,p}$ denote the set of valid selected episodes for task $j$ in stage $p$, and let $\mathcal{J}_d$ denote the set of tasks in dimension $d$. We first average native scores within each task and then aggregate across tasks and dimensions using the benchmark's equal-dimension weighting:
\begin{equation}
 S_{j,p}=\frac{100}{|V_{j,p}|}\sum_{k\in V_{j,p}}s_{j,k},\qquad
 S_{d,p}=\frac{1}{|\mathcal{J}_d|}\sum_{j\in\mathcal{J}_d}S_{j,p},\qquad
 S_p=\frac15\sum_{d=1}^{5}S_{d,p}.
 \label{eq:metric}
\end{equation}
For Generalization, the standard and randomized split means receive equal weight. Since each split contains 12 entries, this is equivalent to averaging all 24 task scores. We compute success rates using the same aggregation procedure applied to explicit native Boolean success flags. Score and success capture different aspects of performance: partial progress may receive a nonzero score without meeting the task's completion criteria.

\paragraph{Reference methods.}
We compare against the three highest-scoring published agent entries, HexaShell-Preview~\citep{gao2026hexashellharnessexploringlimits}, WorldScape-R$^2$SI~\citep{manifold2026r2si}, and EmbodiedEvo~\citep{li2026embodiedevo}; the leading WAM entry, VPP2~\citep{guo2026videopredictionpolicy2}; PhysicalRSI~\citep{physicalrsi2026} as a related adaptation approach; and GPT-6-Astra as a general-purpose agent reference. All reference scores are obtained from the archived public leaderboard \citep{robodojo2026leaderboard}. Although these methods share the same aggregation convention, their interaction budgets, run-selection procedures, and configurations are not matched. The comparisons therefore contextualize overall performance but do not isolate the effects of individual components.

\begin{table}[t]
  \centering
  \caption{\RDCaptionTitle{Overall performance on RoboDojo-Sim.} Scores and native success rates follow the benchmark's five-dimension aggregation across 54 task entries. Reference values are published leaderboard results, while RoboDoctor values summarize selected local outcomes. Evaluation protocols and interaction budgets are not matched.}
  \label{tab:overall}
  \RDTableSetup
  \begin{tabular*}{\linewidth}{@{\extracolsep{\fill}}llrrr@{}}
    \toprule
    \RDTableHeading{Method} & \RDTableHeading{Source} & \RDTableHeading{Score $\uparrow$} & \RDTableHeading{Success (\%) $\uparrow$} & \RDTableHeading{Episodes}\\
    \midrule
    HexaShell-Preview & Published & 53.18 & 46.85 & --- \\
WorldScape-R$^2$SI & Published & 46.58 & 39.35 & --- \\
EmbodiedEvo & Published & 45.08 & 38.95 & --- \\
VPP2 & Published & 39.26 & 32.26 & --- \\
PhysicalRSI & Published & 36.27 & 31.38 & --- \\
GPT-6-Astra & Published & 28.97 & 22.48 & --- \\
\midrule
RoboDoctor (Warmup) & Local archive & 49.74 & 40.83 & 162/162 \\
\RDTableEmphasis{RoboDoctor (Refinement)} & \RDTableEmphasis{Local archive} & \RDTableEmphasis{57.10} & \RDTableEmphasis{49.44} & \RDTableEmphasis{301/324} \\
    \bottomrule
  \end{tabular*}
\end{table}

\subsection{Aggregate Performance and Capability Profile}

\begin{figure}[h]
  \centering
  \includegraphics[width=\linewidth]{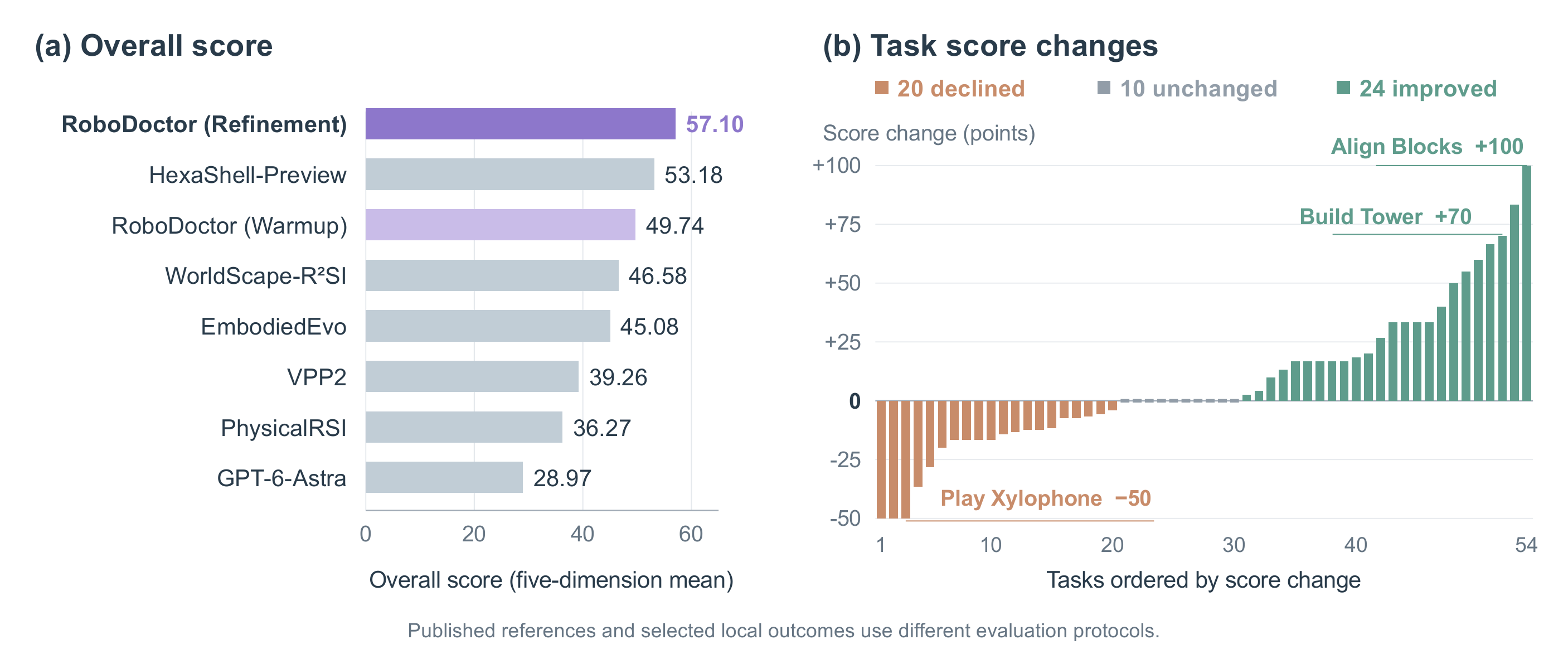}
  \caption{\RDCaptionTitle{Aggregate performance and variation across tasks.} (a) Overall scores for RoboDoctor during Warmup and Refinement and for published reference methods, using the equally weighted mean of five capability scores.
(b) Changes in task-level mean score from Warmup to Refinement across all 54 entries, ordered from the largest decrease to the largest increase: 24 improve, 20 decline, and 10 remain unchanged. Annotations highlight representative gains and a regression. RoboDoctor scores summarize selected local outcomes; reference methods use different evaluation protocols, so the comparison does not isolate the effects of individual components.}
  \label{fig:results}
\end{figure}

The overall score increases from \WarmupScore{} in Warmup to \RefineScore{} in Refinement, corresponding to a gain of \ScoreGain{} points. Figure~\ref{fig:results}(a) places these stage-level scores alongside the published reference results. The native success rate increases from \WarmupSuccess\% to \RefineSuccess\%, a gain of \SuccessGain{} percentage points (Table~\ref{tab:overall}). HexaShell-Preview leads the archived public leaderboard with a score of 53.18 and a success rate of 46.85\%. Refinement exceeds these values by 3.92 points and 2.59 percentage points, respectively, whereas Warmup remains below them. The Refinement score also exceeds those of WorldScape-R$^2$SI, EmbodiedEvo, and VPP2 by 10.52, 12.02, and 17.84 points, respectively. These differences should be interpreted in light of the protocol differences described above.

Improvements vary across capability dimensions (Table~\ref{tab:dimensions}). Generalization and Open improve by 9.48 and 21.35 points, respectively, while Precision and Memory improve by 4.06 and 3.61 points. The gains thus extend across dimensions involving instruction following and precise physical execution. Precision reaches 47.60, compared with 43.03 for VPP2 and 12.65 for GPT-6-Astra in the reference snapshot. However, the Memory score of 77.50 remains below HexaShell-Preview's 84.53 and EmbodiedEvo's 79.40; the Open score of 50.52 also trails HexaShell-Preview's 57.97. Long-Horizon performance decreases from 54.58 to 52.92. The aggregate improvement therefore coexists with persistent limitations in extended task execution. Moreover, the benchmark's Memory dimension evaluates a designated set of tasks and does not independently measure the contribution of our experience mechanism.

\begin{table}[t]
  \centering
  \caption{\RDCaptionTitle{Performance across capability dimensions.} Gen. denotes the equally weighted mean of the standard and randomized Generalization scores. The overall score is the equally weighted mean of the five capability scores. Reference rows report published values; the final two rows summarize selected local outcomes.}
  \label{tab:dimensions}
  \RDTableSetup
  \begin{tabular*}{\linewidth}{@{\extracolsep{\fill}}lrrrrr@{}}
    \toprule
    \RDTableHeading{Method} & \RDTableHeading{Gen.} & \RDTableHeading{Precision} & \RDTableHeading{Long-Hor.} & \RDTableHeading{Memory} & \RDTableHeading{Open}\\
    \midrule
    HexaShell-Preview & 39.22 & 41.25 & 42.91 & 84.53 & 57.97 \\
WorldScape-R$^2$SI & 42.95 & 23.51 & 51.48 & 65.78 & 49.16 \\
EmbodiedEvo & 35.22 & 25.15 & 33.83 & 79.40 & 51.81 \\
VPP2 & 36.38 & 43.03 & 56.28 & 56.01 & 4.61 \\
PhysicalRSI & 21.30 & 38.05 & 46.37 & 46.74 & 28.88 \\
GPT-6-Astra & 33.36 & 12.65 & 21.45 & 43.04 & 34.36 \\
\midrule
RoboDoctor (Warmup) & 47.50 & 43.54 & 54.58 & 73.89 & 29.17 \\
\RDTableEmphasis{RoboDoctor (Refinement)} & \RDTableEmphasis{56.98} & \RDTableEmphasis{47.60} & \RDTableEmphasis{52.92} & \RDTableEmphasis{77.50} & \RDTableEmphasis{50.52} \\
    \bottomrule
  \end{tabular*}
  \par\vspace{4pt}
  {\small\raggedright Task entries: 24 Generalization; 8 Precision; 8 Long-Horizon; 6 Memory; 8 Open.\par}
\end{table}

\subsection{Task-Level Variation and Interpretation}

Figure~\ref{fig:results}(b) orders the 54 task entries by the change in their mean score from Warmup to Refinement, revealing substantial variation underlying the aggregate improvement: 24 tasks improve, 20 decline, and 10 remain unchanged. The annotated gains include Align Blocks, which improves from 0 to 100, and Build Tower, which improves from 30 to 100; Insert Tubes also improves from 40 to 80. At the negative end, Play Xylophone and Match and Pick from Conveyor both decline from 100 to 50. These heterogeneous changes motivate a closer examination of the physical decisions associated with individual outcomes.

The stage comparison jointly reflects additional interaction, experience revision, route selection, and differences among source runs. It therefore does not isolate the contribution of any individual component, including memory. In the following section, we analyze recorded cases to characterize how the framework modifies physical execution. Establishing the causal contribution of each mechanism requires matched experiments with controlled routes, seeds, and interaction budgets.

\section{Analysis of Execution Changes}
\label{sec:cases}

Figure~\ref{fig:cases} illustrates execution changes at the three revision levels through the initial situation, the revised interaction, and the observed result. Align Blocks illustrates a change in task strategy; Pour Liquid into Cup illustrates a targeted correction that preserves an existing grasp; and Build Tower illustrates a revised grasp verified before placement. We analyze the visible transitions in these examples to distinguish the role of each revision level.

\subsection{Harness: Revising the Interaction Strategy}

Figure~\ref{fig:cases}(a) contrasts an unaligned configuration from Warmup with supported pushing in a later \Direct{} episode. In the revised interaction, the set square remains supported on the table while the two grippers contact spatially separated points along its edge, highlighted by circles. The final view shows the blocks aligned along the tool.

The change concerns how the tool is used to organize the task: the later strategy combines table support with coordinated contact from both grippers. This illustrates the role of Harness RSI in allocating control to an agent-led interaction strategy when the preceding execution has left the task unresolved.

\subsection{Intervention: Correcting Alignment While Preserving the Grasp}

Figure~\ref{fig:cases}(b) shows a bottle held with its mouth laterally offset from the cup. The Doctor applies the illustrated one-step lateral correction while maintaining the grasp. The policy then resumes tilting and pours into the cup before withdrawing the bottle.

This sequence illustrates selective assistance at the level of alignment. The correction preserves the grasp that supports the ongoing manipulation, allowing the policy to continue the pouring motion. The case highlights the role of Intervention RSI in identifying a local mismatch and correcting it while retaining useful policy behavior.

\subsection{Operation: Revising and Verifying the Grasp}

\begin{figure}[t]
  \centering
  \includegraphics[width=\linewidth]{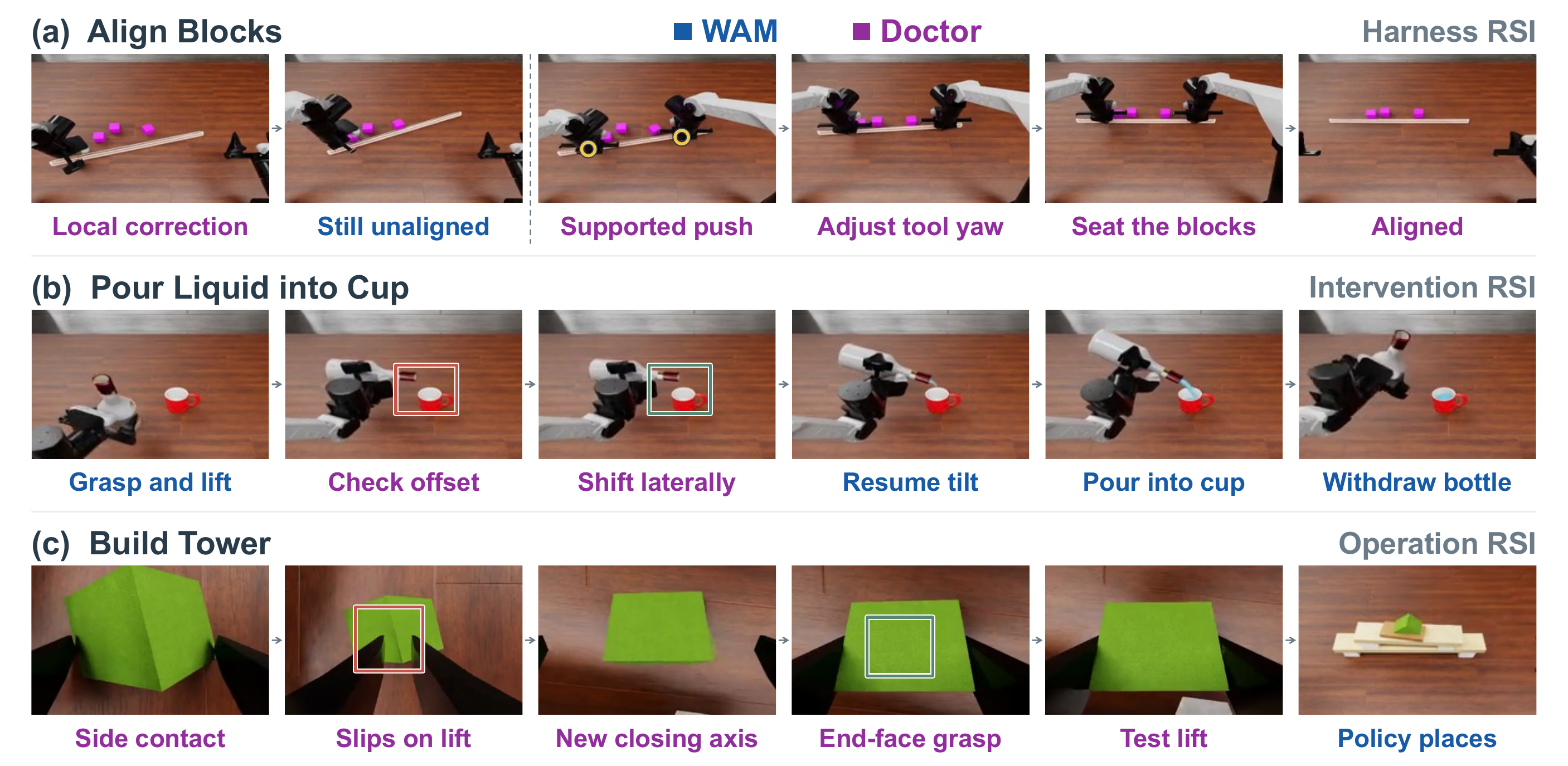}
  \caption{\RDCaptionTitle{Recorded execution changes across three RSI levels.} (a) Harness RSI: an unresolved Warmup attempt is contrasted with supported tool pushing in a later DIRECT episode; the dashed divider separates episodes. (b) Intervention RSI: the Doctor corrects bottle–cup alignment while preserving the grasp, after which the WAM pours and withdraws the bottle. (c) Operation RSI: the Doctor revises a slipped grasp and verifies retention with a test lift before WAM-led placement. Blue labels indicate WAM actions or outcomes; purple labels indicate Doctor actions, outcomes, or diagnosis. Frames show selected moments within each episode.}
  \label{fig:cases}
\end{figure}

Figure~\ref{fig:cases}(c) shows a progression from slippage to an end-face grasp, followed by a test lift and placement. The revised contact changes how the gripper holds the green block. During the test lift, the block remains between the fingers; the final view shows it placed on the supporting structure.

The test lift provides an intermediate check of the revised contact before placement. This illustrates the role of Operation RSI in modifying the physical method and assessing its observed effect. Grasp retention during this check supports proceeding with the manipulation, while the subsequent placement provides separate evidence of progress toward the task objective.

\section{Discussion and Limitations}
\paragraph{Evidence determines what can be reused.}
The three revision targets distinguish failures that require different responses: reallocate control, change an intervention decision, or revise the contact method. Their connection through experience is useful only if the retained lesson accurately reflects execution. A plausible command or achieved robot pose does not certify stable contact, and visible progress need not satisfy the native task predicate. Schema validation makes records inspectable but does not establish their truth. Perceptual errors, incorrect diagnoses, and overgeneralized lessons can therefore persist across decisions.

\paragraph{Mechanism evidence and causal evidence.}
The cases demonstrate strategy changes, selective correction, and feedback-driven operation revision. The aggregate results characterize the selected runs, including regressions. Neither alone establishes the causal benefit of cross-episode experience. Retrospective selection, unequal coverage, and differing source histories limit attribution. A prospective study should compare persistent and reset experience under matched controllers, configurations, seeds, and interaction budgets. The same design should compare fixed \Hybrid{}, fixed \Direct{}, and Warmup-based routing.

\paragraph{Cost and generality.}
Training-free adaptation still consumes agent inference, policy calls, and environment interaction. Short execution horizons can improve observability while increasing latency; the present archive lacks a consolidated cost analysis. All results are in simulation, and transfer to other policies or physical robots is unmeasured. Bounded actions limit motion before renewed observation, but do not provide collision avoidance or a formal safety guarantee.

\section{Conclusion}
\method{} studies how existing model capabilities can be adapted to a task through physical interaction. With policy and agent weights fixed, it connects control allocation, selective intervention, and operation revision through experience that retains both a method and its supporting evidence. This makes execution a source of adaptation: observed effects inform what to preserve, what to revise, and what can guide a later attempt. The RoboDojo-Sim archive shows a higher aggregate Refinement score alongside concrete strategy changes and local corrections, with regressions that expose remaining limitations. The next experimental priority is to measure when revised experience improves subsequent execution under matched configurations and interaction budgets.

\bibliography{references}
\bibliographystyle{iclr2027_conference}
\clearpage
\appendix
\section{Runtime Contracts and External Experience}
\label{app:runtime}
The implementation separates policy proposal from physical execution. \code{policy\_rollout} validates a finite $H\times14$ array and gripper values in $[0,1]$, requests a robot-only FK preview, and returns with zero physical steps. \code{policy\_continue} requires the current proposal identifier and executes an unmodified prefix, capped at the controller's execution limit. Direct-tool interventions in \Hybrid{} require an existing proposal and a step budget from one to five. A new observation clears the old proposal. These host-side checks enforce the protocol independently of whether the agent's diagnosis is correct.

The current RSI controller caps released prefixes at 16 steps, even if a backend advertises a longer execution horizon. The bundled PrismWAM preview configuration advertises 32-action predictions and a 24-step backend horizon; the host's 16-step cap determines what a single \Hybrid{} release can execute. This distinction prevents backend defaults from being mistaken for the applied controller limit. The current preview backend uses four past RGB frames at stride four plus the current frame; early missing frames repeat the earliest observation. These implementation settings are not evidence that every archived run used an identical checkpoint or history configuration.

\begin{table}[h]
\caption{\RDCaptionTitle{Evidence contracts for grounded action.} These checks distinguish levels of support; none replaces the final native task outcome.}
\label{tab:evidence}
\RDTableSetup
\renewcommand{\arraystretch}{1.1}
\begin{tabularx}{\linewidth}{@{}L{.23\linewidth}YY@{}}
\toprule
\RDTableHeading{Evidence} & \RDTableHeading{What it supports} & \RDTableHeading{What it does not establish}\\
\midrule
Policy proposal / FK preview & Intended joint targets and predicted robot poses & Actual tracking, object motion, contact, or task success\\
\addlinespace[2pt]
Measured end-effector pose & Achieved robot pose and target error & Stable grasp or correct placement\\
\addlinespace[2pt]
Fresh RGB-D interaction & Observed contact, support, transfer, or visible failure & Hidden quantitative predicates or full-task completion\\
\addlinespace[2pt]
Native outcome & Evaluator-defined score, validity, and success & The causal benefit of one intervention or memory entry\\
\bottomrule
\end{tabularx}
\end{table}

\paragraph{\textmd{\Hybrid{}} memory fields.}
The experience document contains a schema identifier, version, task, latest iteration, and two blocks. Diagnosis entries retain observation references, RGB-D and robot state, action status (proposed versus executed), measured versus previewed FK, failure or success evidence, and the associated intervention when present. Operation entries retain timing, tools, technique, stop condition, observed outcome, and any referenced policy method and its adaptation. An object-interaction description can include the target object, acting arm, interaction type, contact region, approach direction, end-effector orientation, gripper sequence, motion sequence, observed success evidence, and applicability conditions. Unknown geometry remains qualitative or unknown. These fields make a lesson inspectable; they do not mechanically prove the truth of its natural-language interpretation.

\paragraph{\textmd{\Direct{}} task experience.}
The \Direct{} controller keeps a current plan in \code{TODO.md} and detailed evidence in \code{NOTES.md}. Its post-episode consolidation is action-disabled. The host appends the native outcome and distinguishes a completed model summary from an evidence fallback. The next episode receives prior lessons as reference material, while current progress, object poses, and absolute coordinates must be reconstructed from its own observations. This differs from simply continuing the previous episode's task state.

\paragraph{Tool and guidance scope.}
Grounding maps native-image pixels to observed surface points. Wrist depth is measured along the camera axis and is re-queried after movement. Both end-effector and joint-space commands specify the waiting arm explicitly. Gripper commands take effect together with motion from the first step; the supplied guidance therefore separates final alignment from closure when necessary. \Direct{} can use up to 150 steps for coarse motion on a confidently clear route, whereas \Hybrid{} interventions retain the hard five-step cap. Guidance about contact regions, test lifts, arm roles, and viewpoint changes is supplied before interaction and should not be presented as knowledge newly discovered in every run.

\Needspace{12\baselineskip}
\section{Complete Per-Task Results}
\begingroup
\RDTableSetup
\renewcommand{\arraystretch}{1.18}
\urlstyle{same}
\setlength{\LTleft}{0pt}
\setlength{\LTright}{0pt}
\setlength{\LTcapwidth}{\linewidth}
\setlength{\RDTableContentWidth}{\dimexpr\linewidth-12\tabcolsep\relax}
\begin{longtable}{@{}L{.42\RDTableContentWidth}L{.09\RDTableContentWidth}R{.09\RDTableContentWidth}R{.09\RDTableContentWidth}R{.11\RDTableContentWidth}R{.11\RDTableContentWidth}R{.09\RDTableContentWidth}@{}}
\caption{\RDCaptionTitle{All 54 task entries.} W: Warmup score; R: Refinement score; $\Delta=\mathrm R-\mathrm W$; SR: Refinement success rate (\%); $n$: valid Refinement episodes out of six nominal slots. All tasks have three Warmup outcomes. Route describes the selected Refinement records; Mixed identifies retrospective combination of both modes. Native task IDs are preserved.}\label{tab:tasks}\\
\toprule
\RDTableHeading{Task} & \RDTableHeading{Route} & \RDTableHeading{W} & \RDTableHeading{R} & \RDTableHeading{$\Delta$} & \RDTableHeading{SR} & \RDTableHeading{$n$}\\
\midrule
\endfirsthead
\multicolumn{7}{l}{Table~\ref{tab:tasks} continued}\\
\toprule
\RDTableHeading{Task} & \RDTableHeading{Route} & \RDTableHeading{W} & \RDTableHeading{R} & \RDTableHeading{$\Delta$} & \RDTableHeading{SR} & \RDTableHeading{$n$}\\
\midrule
\endhead
\midrule
\multicolumn{7}{r}{Continued on next page}\\
\endfoot
\bottomrule
\endlastfoot
\nolinkurl{align_blocks} & Direct & 0.00 & 100.00 & +100.00 & 100.00 & 6/6 \\
\nolinkurl{arrange_largest_number} & Hybrid & 76.67 & 63.33 & -13.33 & 50.00 & 6/6 \\
\nolinkurl{arrange_largest_number_random} & Hybrid & 48.33 & 61.67 & +13.33 & 50.00 & 6/6 \\
\nolinkurl{build_tower} & Hybrid & 30.00 & 100.00 & +70.00 & 100.00 & 3/6 \\
\nolinkurl{classify_objects} & Hybrid & 100.00 & 80.00 & -20.00 & 66.67 & 6/6 \\
\nolinkurl{classify_objects_by_language} & Direct & 26.67 & 43.33 & +16.67 & 16.67 & 6/6 \\
\nolinkurl{cover_blocks} & Hybrid & 100.00 & 88.33 & -11.67 & 83.33 & 6/6 \\
\nolinkurl{deposit_coin} & Direct & 46.67 & 10.00 & -36.67 & 0.00 & 6/6 \\
\nolinkurl{fasten_screws} & Hybrid & 83.33 & 66.67 & -16.67 & 33.33 & 6/6 \\
\nolinkurl{fill_egg_holder} & Direct & 3.33 & 5.83 & +2.50 & 0.00 & 6/6 \\
\nolinkurl{fill_pen_holder} & Hybrid & 35.00 & 90.00 & +55.00 & 83.33 & 6/6 \\
\nolinkurl{fold_clothes} & Hybrid & 40.00 & 100.00 & +60.00 & 100.00 & 6/6 \\
\nolinkurl{fold_clothes_random} & Hybrid & 100.00 & 100.00 & +0.00 & 100.00 & 6/6 \\
\nolinkurl{general_pickup} & Hybrid & 100.00 & 100.00 & +0.00 & 100.00 & 6/6 \\
\nolinkurl{hang_mugs} & Hybrid & 43.33 & 70.00 & +26.67 & 50.00 & 6/6 \\
\nolinkurl{hang_mugs_random} & Direct & 10.00 & 2.50 & -7.50 & 0.00 & 6/6 \\
\nolinkurl{imitate_sorting_sequence} & Hybrid & 76.67 & 76.67 & +0.00 & 66.67 & 3/6 \\
\nolinkurl{insert_key} & Direct & 15.00 & 7.50 & -7.50 & 0.00 & 6/6 \\
\nolinkurl{insert_tubes} & Hybrid & 40.00 & 80.00 & +40.00 & 66.67 & 3/6 \\
\nolinkurl{make_kong} & Hybrid & 66.67 & 16.67 & -50.00 & 16.67 & 6/6 \\
\nolinkurl{make_toast} & Hybrid & 66.67 & 54.17 & -12.50 & 50.00 & 6/6 \\
\nolinkurl{make_toast_random} & Hybrid & 41.67 & 37.50 & -4.17 & 33.33 & 6/6 \\
\nolinkurl{match_and_pick_from_conveyor} & Hybrid & 100.00 & 50.00 & -50.00 & 50.00 & 6/6 \\
\nolinkurl{organize_table} & Hybrid & 66.67 & 54.17 & -12.50 & 0.00 & 6/6 \\
\nolinkurl{pack_objects_into_box} & Direct & 20.00 & 40.00 & +20.00 & 33.33 & 3/6 \\
\nolinkurl{pack_objects_into_box_random} & Hybrid & 53.33 & 47.50 & -5.83 & 33.33 & 6/6 \\
\nolinkurl{pick_from_conveyor_by_image} & Direct & 0.00 & 0.00 & +0.00 & 0.00 & 6/6 \\
\nolinkurl{play_Xylophone} & Hybrid & 100.00 & 50.00 & -50.00 & 50.00 & 6/6 \\
\nolinkurl{play_stacking_toy} & Direct & 6.67 & 25.00 & +18.33 & 0.00 & 4/6 \\
\nolinkurl{play_tic_tac_toe} & Hybrid & 58.33 & 51.67 & -6.67 & 16.67 & 6/6 \\
\nolinkurl{plug_in_charger} & Direct & 0.00 & 33.33 & +33.33 & 33.33 & 6/6 \\
\nolinkurl{pour_balls_into_vase} & Hybrid & 33.33 & 33.33 & +0.00 & 33.33 & 6/6 \\
\nolinkurl{pour_by_language} & Direct & 0.00 & 16.67 & +16.67 & 16.67 & 6/6 \\
\nolinkurl{pour_liquid_into_cup} & Hybrid & 66.67 & 66.67 & +0.00 & 66.67 & 3/6 \\
\nolinkurl{pour_liquid_into_cup_random} & Direct & 0.00 & 33.33 & +33.33 & 33.33 & 6/6 \\
\nolinkurl{press_by_number} & Mixed & 100.00 & 83.33 & -16.67 & 83.33 & 6/6 \\
\nolinkurl{push_T} & Direct & 0.00 & 50.00 & +50.00 & 50.00 & 6/6 \\
\nolinkurl{push_T_random} & Direct & 0.00 & 83.33 & +83.33 & 83.33 & 6/6 \\
\nolinkurl{put_bottles_into_dustbin} & Hybrid & 100.00 & 100.00 & +0.00 & 100.00 & 3/6 \\
\nolinkurl{solve_equation} & Hybrid & 66.67 & 66.67 & +0.00 & 66.67 & 6/6 \\
\nolinkurl{sort_nesting_dolls_by_size} & Hybrid & 66.67 & 50.00 & -16.67 & 50.00 & 6/6 \\
\nolinkurl{sort_nesting_dolls_by_size_random} & Direct & 0.00 & 16.67 & +16.67 & 16.67 & 6/6 \\
\nolinkurl{stack_blocks} & Hybrid & 100.00 & 85.83 & -14.17 & 83.33 & 6/6 \\
\nolinkurl{stack_blocks_by_language} & Hybrid & 40.00 & 73.33 & +33.33 & 66.67 & 6/6 \\
\nolinkurl{stack_blocks_random} & Hybrid & 100.00 & 100.00 & +0.00 & 100.00 & 6/6 \\
\nolinkurl{stack_bowls} & Hybrid & 100.00 & 100.00 & +0.00 & 100.00 & 6/6 \\
\nolinkurl{stack_bowls_random} & Hybrid & 100.00 & 71.67 & -28.33 & 66.67 & 6/6 \\
\nolinkurl{store_laptop_and_headphones} & Hybrid & 46.67 & 56.67 & +10.00 & 33.33 & 6/6 \\
\nolinkurl{store_laptop_and_headphones_random} & Hybrid & 60.00 & 43.33 & -16.67 & 16.67 & 6/6 \\
\nolinkurl{store_tools_in_toolbox} & Direct & 0.00 & 4.17 & +4.17 & 0.00 & 6/6 \\
\nolinkurl{swap_T} & Hybrid & 33.33 & 66.67 & +33.33 & 66.67 & 3/6 \\
\nolinkurl{swap_blocks} & Hybrid & 33.33 & 100.00 & +66.67 & 100.00 & 6/6 \\
\nolinkurl{sweep_blocks} & Direct & 0.00 & 16.67 & +16.67 & 16.67 & 6/6 \\
\nolinkurl{sweep_blocks_random} & Direct & 0.00 & 16.67 & +16.67 & 16.67 & 6/6 \\
\end{longtable}
\endgroup

\section{Case Evidence and Interpretation}
\label{app:cases}
Figure~\ref{fig:cases} presents selected recorded views of execution changes and their observed effects. Figure~\ref{fig:overview} combines a conceptual illustration of the three revision levels with capability scores and examples of alignment correction and grasp revision. Figure~\ref{fig:architecture} depicts control allocation, intervention, grounded operation, and experience update, using pouring to illustrate how a local correction supports subsequent policy motion. The case examples illustrate distinct interactions rather than a single continuous episode or measured transfer of experience between tasks. Experience fields and update rules are specified in Section~\ref{sec:experience} and Appendix~\ref{app:runtime}.

\begin{table}[h]
\caption{\RDCaptionTitle{Visual evidence in Figure~\ref{fig:cases}.} Each example connects a revised interaction to its depicted outcome.}
\RDTableSetup
\begin{tabularx}{\linewidth}{@{}L{.23\linewidth}L{.25\linewidth}Y@{}}
\toprule
\RDTableHeading{Task / panel} & \RDTableHeading{Visible evidence} & \RDTableHeading{Supported interpretation}\\
\midrule
Align Blocks, Figure~\ref{fig:cases}(a) & Unaligned blocks; two-point supported pushing; aligned blocks & The later \Direct{} episode uses a supported pushing strategy to align the blocks.\\
\addlinespace[4pt]
Pour Liquid into Cup, Figure~\ref{fig:cases}(b) & Lateral offset; correction with the grasp retained; pouring & Local alignment is corrected while preserving the grasp, followed by policy-driven pouring.\\
\addlinespace[4pt]
Build Tower, Figure~\ref{fig:cases}(c) & Slippage; end-face grasp; test lift; placement & Revised contact is checked through a lift before the block is placed.\\
\bottomrule
\end{tabularx}
\end{table}

The cases illustrate local changes in physical execution. Liquid entering the cup and the block remaining held during a test lift provide evidence of the depicted interactions; they do not independently establish evaluator-defined task success or the benefit of experience reuse across episodes.

\end{document}